\documentclass[runningheads]{llncs}
\usepackage{times,multirow,booktabs,graphicx,float,array,subcaption,caption,pifont,amsmath,xcolor,hyperref,bbding}
\usepackage[most]{tcolorbox}
\usepackage{svg}
\usepackage{multirow}
\usepackage[table]{xcolor} 
\usepackage{tabularx}
\newcolumntype{Y}{>{\centering\arraybackslash}X} %
\usepackage[framemethod=tikz]{mdframed}
\usepackage[T1]{fontenc}
\usepackage{graphicx}
\usepackage{amsmath}
\usepackage{amssymb} 
\definecolor{lightgreen}{HTML}{E0F2E9}
\definecolor{lightred}{HTML}{FDE0E0}
\usepackage{color}

\usepackage{xspace}

\newcommand{\method}{\texttt{GlassMol}\xspace}

\begin{document}
\title{\method: Interpretable Molecular Property Prediction with Concept Bottleneck Models}
%
%
\author{Oscar Rivera\Envelope\inst{1}\orcidID{0009-0005-5742-1740} \and Ziqing Wang\inst{1}\orcidID{0009-0004-8940-0461} \and Matthieu Dagommer \inst{2} \orcidID{0009-0006-3260-6455} \and Abhishek Pandey \inst{2}\orcidID{0009-0008-7905-5712} \and Kaize Ding\inst{1}\orcidID{0000-0001-6684-6752}}
%
\authorrunning{O. Rivera, Z. Wang, et al.}
%

\institute{Northwestern University, Evanston, IL 60201, USA
\\\email{oscarrivera2028@u.northwestern.edu}\\
\and
AbbVie, Chicago, IL 60064, USA \\}
\titlerunning{\method}
\maketitle              
\vspace{-6mm}
\begin{abstract}
Machine learning accelerates molecular property prediction, yet state-of-the-art Large Language Models and Graph Neural Networks operate as black boxes. In drug discovery, where safety is critical, this opacity risks masking false correlations and excluding human expertise. Existing interpretability methods suffer from the \textbf{effectiveness-trustworthiness} trade-off: explanations may fail to reflect a model's true reasoning, degrade performance, or lack domain grounding. Concept Bottleneck Models (CBMs) offer a solution by projecting inputs to human-interpretable concepts before readout, ensuring that explanations are inherently faithful to the decision process. However, adapting CBMs to chemistry faces three challenges: the Relevance Gap (selecting task-relevant concepts from a large descriptor space), the Annotation Gap (obtaining concept supervision for molecular data), and the Capacity Gap (degrading performance due to bottleneck constraints). We introduce \method, a model-agnostic CBM that addresses these gaps through automated concept curation and LLM-guided concept selection. Experiments across thirteen benchmarks demonstrate that \method generally matches or exceeds black-box baselines, suggesting that interpretability does not sacrifice performance and challenging the commonly assumed trade-off. Code is available at \url{https://github.com/walleio/GlassMol}.

\end{abstract}
\vspace{-8mm}
\section{Introduction}
\vspace{-2mm}
Machine learning-enabled molecular property prediction seeks to leverage the high-throughput nature of computation to identify promising areas of research for wet-lab chemists, increasing the efficiency of the entire research pipeline \cite{designofefficient,synthesis,denovo}. By learning from vast chemical databases where exhaustive experimental testing is prohibitively expensive, ML models prioritize promising compounds for wet-lab validation, reducing both cost and development timelines. This computational approach has already demonstrated tangible impact in identifying novel antibiotics, optimizing molecular properties, and enabling de novo drug design \cite{designofefficient,synthesis,polo,mlindrugdisc}.
\vspace{0.1cm}
\\
Among the most successful approaches are Graph Neural Networks (GNNs) \cite{rossjerret,ju2025textm2llmmultiviewmolecularrepresentation,chenself,ding2022data} and Large Language Models (LLMs) \cite{wang2025survey,wen2024alignab}. By design, GNNs rely on graph-structured data, enabling message-passing between nodes (i.e., atoms), and are thus well suited for processing intrinsically graph-like molecular information \cite{naturedrugs}. Though LLMs fail to exploit graph structure, they have demonstrated strong encoding and reasoning capabilities for natural-language molecular tasks \cite{llmgnn}. Both paradigms achieve strong predictive performance, yet both operate as black boxes: GNNs produce high-dimensional graph embeddings that resist interpretation, while LLMs function through billions of opaque parameters \cite{sruv,yuan2021explainabilitygraphneuralnetworks}. In drug discovery, where regulatory approval and patient safety demand rigorous justification, such opacity poses significant challenges.
\vspace{0.1cm}
\\
The need for interpretability extends beyond regulatory compliance to fundamental questions of scientific validity. Without understanding \textit{why} a model predicts toxicity, researchers cannot verify whether predictions reflect genuine structure-activity relationships or spurious correlations. A model might correctly flag a molecule as hepatotoxic for reasons unrelated to its actual mechanism, making it difficult to guide lead optimization. Existing interpretability methods, including attention visualization, gradient-based attribution, and subgraph exploration, attempt to address this gap \cite{yuan2021explainabilitygraphneuralnetworks,ye2022unreliabilityexplanationsfewshotprompting}. However, these post-hoc approaches face a fundamental \textbf{effectiveness-trustworthiness} trade-off: explanations may degrade predictive accuracy, may not faithfully reflect actual model reasoning, or may lack grounding in chemical knowledge \cite{radhakrishnan2023questiondecompositionimprovesfaithfulness}.
\vspace{0.1cm}
\\
Concept Bottleneck Models (CBMs) offer a principled alternative by embedding interpretability directly into model architecture \cite{cbmoriginal,belinkov2019analysis,madsen2022post}. A CBM first predicts human-understan\-dable intermediate concepts, then uses \textit{exclusively} these concepts for final predictions. This design ensures that explanations are inherently faithful to the decision process, as every prediction must pass through the interpretable concept layer. However, adapting CBMs to molecular property prediction presents domain-specific challenges:
\vspace{-2mm}
\begin{enumerate}
\item[\ding{202}] \textit{How to select concepts? (Relevance Gap)}: Unlike computer vision, where concepts are intuitive (e.g., wing color), the chemical descriptor space contains hundreds of physicochemical properties. Identifying the specific subset relevant to a downstream task (e.g., toxicity) is non-trivial and often exceeds manual expert capacity.
\item[\ding{203}] \textit{Where is the data? (Annotation Gap)}: Standard molecular datasets provide only structures and final labels. Training a CBM requires supervision on intermediate concepts, which are typically absent, creating a barrier to entry.
\item[\ding{204}] \textit{Are performance and transparency mutually exclusive? (Capacity Gap)}: There is a common concern that imposing a bottleneck of human-interpretable concepts may constrain the model's expressivity, leading to inferior performance compared to black-box end-to-end models.
\end{enumerate}
\vspace{-2mm}
\indent To address these challenges, we propose \method, a model-agnostic adaptation of CBMs to molecular property prediction. Although formally model-agnostic, we experimentally verify \method using GNN and LLM backbones because of their relevance and suitability for chemical tasks \cite{3dllmgnn,fung2021benchmarking,rossjerret}. \method bridges the Annotation Gap by computing chemical descriptors via RDKit, addresses the Relevance Gap through LLM-guided concept selection, and empirically demonstrates that interpretability need not sacrifice performance. In summary, this paper makes the following contributions:
\vspace{-2mm}
\begin{itemize}
\item \textbf{Method.} A model-agnostic CBM framework and corresponding open-source code applicable to molecular property prediction tasks with continuous concept labels.
\item \textbf{Evaluation.} A systematic evaluation on thirteen benchmark datasets demonstrating competitive or superior performance compared to black-box baselines, empirically addressing the Capacity Gap.
\item \textbf{Verification.} A grounded validation of interpretability through case studies comparing concept attributions with established structural importance methods.
\end{itemize}
\vspace{-2mm}
\begin{figure}[t!]
\centering
\includegraphics[height=4.7cm]{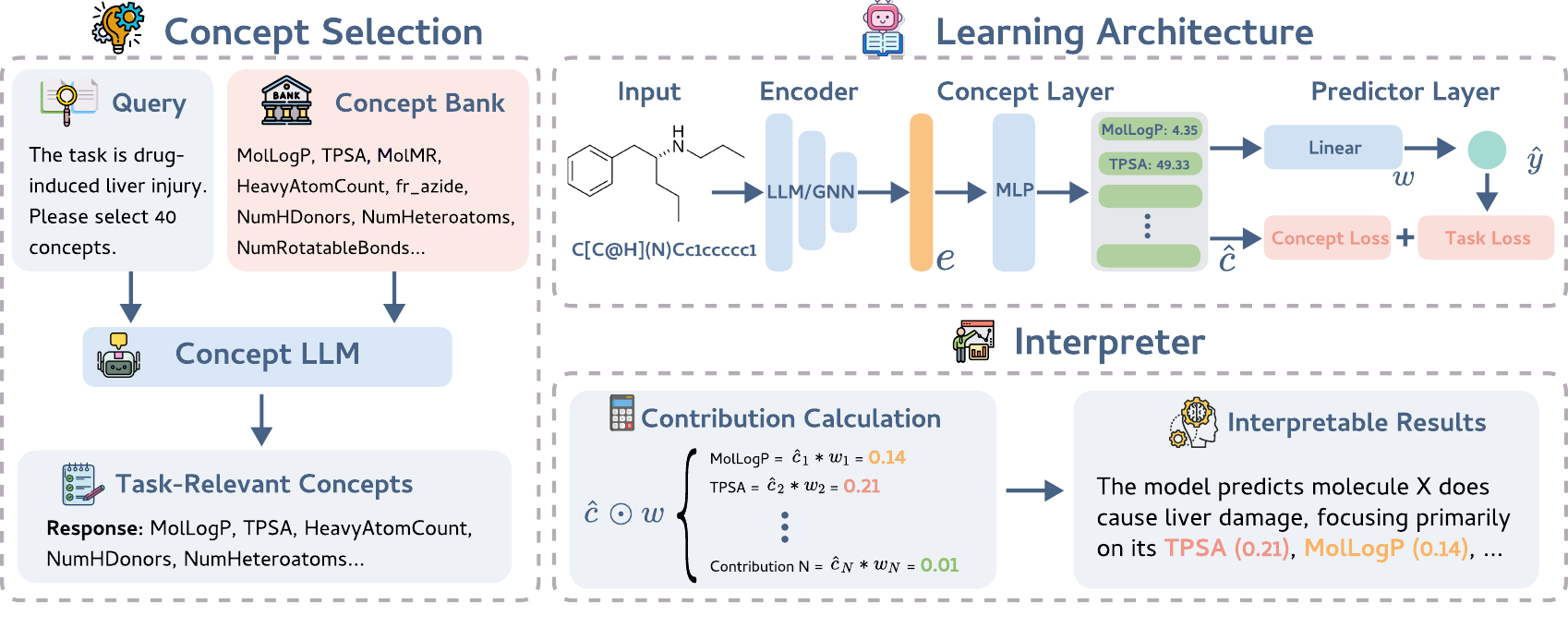}
\vspace{-4mm}
\caption{\textbf{The proposed \method framework}. The inputs to the encoder are the raw data, either a SMILES string or a molecular graph. The embeddings generated by the LLM or GNN are projected onto a set of task-relevant concepts selected by an LLM. The predicted concept values are then passed to a linear layer to produce the final prediction. The importance of each concept is determined by the magnitude of its contribution to the final output, computed as the product of the concept value and its weight.}
\label{fig:1}
\vspace{-0.6cm}
\end{figure}
\vspace{-4mm}
\section{Related Work}
\vspace{-2mm}
\textit{\textbf{Molecular Representation Learning.}} GNNs~\cite{kipf2017semisupervisedclassificationgraphconvolutional} have been extensively applied to 2D and 3D molecular tasks, utilizing supervised and unsupervised approaches to capture topological nuances \cite{hu2025improving,fung2021benchmarking}. LLM-based methods, including chemical pre-training and prompting variations, have performed comparably to their GNN counterparts \cite{rossjerret,ju2025textm2llmmultiviewmolecularrepresentation}. Recent research efforts, moreover, seek to synthesize GNN and LLM embeddings to maximize information expressibility \cite{3dllmgnn,fan2026hierarchical}. Though both learning strategies exhibit state-of-the-art performance on molecular tasks, they are not designed with an interpret\-ability-first approach, limiting collaboration with domain experts \cite{li2025kolmogorov,llmprop}. Indeed, GNNs and LLMs intrinsically lack interpretability: GNNs encode molecular structures into dense vector representations that are difficult to decompose, while LLMs distribute knowledge across billions of parameters without explicit reasoning traces \cite{sruv,yuan2022explainabilitygraphneuralnetworks}.
\smallskip
\\
\noindent\textit{\textbf{Explainability in Drug Discovery.}} To address this opacity, add-on methods often gauge model behavior using data manipulation, input probing, subgraph exploration, and uncertainty quantification. Although not requiring re-training, these post-hoc approaches face the ``trustworthiness'' problem: they approximate the model's reasoning rather than exposing it, potentially learning faulty concepts \cite{tvac,drugsurv,dai2021towards}. In contrast, self-explanatory methods are considered the gold standard for interpretability in chemistry because they are human-intelligible and integrated into the model \cite{qadri2025explainable}. However, existing self-explanatory approaches often synthesize exclusively subgraph-based explanations, remaining difficult for non-experts to interpret \cite{GLAD,subgraph,cao2024group}.
\smallskip
\\
\noindent\textit{\textbf{Concept Bottleneck Models.}} CBMs offer a path toward self-explanation by projecting latent features into a semantic concept space before prediction \cite{cbmoriginal,cbmoriginal2} and have demonstrated their efficacy on computer vision and text-based tasks \cite{kohetal,hatespeech,tan2023interpretingpretrainedlanguagemodels}. Several architectures have been proposed: linear layers projecting from hidden states to predefined concepts \cite{tan2023interpretingpretrainedlanguagemodels}, and neuron-level concept representations \cite{ismail2024conceptbottlenecklanguagemodels}. In the image domain, LLMs have been used to curate concept banks \cite{bank}. However, these approaches have not been systematically adapted to molecular property prediction, where the Relevance Gap and Annotation Gap pose challenges. \method builds upon these foundations while following the guidance of \cite{leakage} to avoid concept prediction inaccuracy.
\vspace{-8mm}
\section{Proposed Approach}
\label{sec:method}

\vspace{-2mm}
In this section, we introduce \method, a model-agnostic framework designed for transparency in molecular property prediction. Though formally model-agnostic, we experimentally verify our approach on LLMs and GNNs because of their relevance and aptitude for chemical tasks \cite{llmprop,3dllmgnn,li2025kolmogorov}. By explicitly aligning high-dimensional molecular embeddings with human-understandable chemical concepts, \method bri\-dges the gap between black-box performance and scientific interpretability.

\vspace{-4mm}
\subsection{Problem Formulation}
\label{subsec:problem}
\vspace{-2mm}

Let $\mathcal{D} = \{(x_i, y_i)\}_{i=1}^N$ denote a molecular dataset, where $x_i \in \mathcal{X}$ is the molecular input (e.g., graph or SMILES) and $y_i \in \mathcal{Y}$ is the target property. 
Standard black-box models learn a direct mapping $F: \mathcal{X} \to \mathcal{Y}$. In contrast, CBMs introduce an intermediate concept space $\mathcal{C} \subseteq \mathbb{R}^K$, decomposing the prediction into a structured process:
\begin{equation}
    \hat{y}_i = h_\psi(\underbrace{g_\phi(f_\theta(x_i))}_{\hat{\mathbf{c}}_i}) = \sigma(\mathbf{w}^\top \hat{\mathbf{c}}_i + b)
\end{equation}
Here, $f_\theta$ extracts latent embeddings, $g_\phi$ projects them to interpretable concept scores $\hat{\mathbf{c}}_i \in \mathbb{R}^K$, and $h_\psi$ makes the final prediction via a linear combination parameterized by weights $\mathbf{w}$. This linearity ensures that the decision logic is transparent.
\smallskip
\\
\noindent\textit{\textbf{Challenges.}} Training such a system requires supervision on the intermediate concepts $\hat{\mathbf{c}}_i$, but molecular datasets lack ground-truth concept annotations $\mathbf{c}^*$, creating an Annotation Gap. Further, the chemical descriptor space is vast, making the identification of a task-relevant subset of $K$ concepts a non-trivial Relevance Gap. \method addresses both gaps by constructing a surrogate ground-truth $\mathbf{c}^*$ via automated domain reasoning.

\vspace{-4mm}
\subsection{Automated Concept Curation via LLM}
\label{subsec:curation}
\vspace{-2mm}

To bridge the Annotation and Relevance Gaps, we propose a two-stage curation pipeline that leverages computational chemistry for precision and LLMs for reasoning.
\smallskip
\\
\noindent\textbf{Stage 1: Ground Truth Generation.} 
First, to address the Annotation Gap, we utilize RDKit \cite{rdkit} as a computational oracle. For every molecule $x_i \in \mathcal{D}$, we compute a comprehensive set of $M=200$ physicochemical descriptors, denoted as the global concept pool $\mathcal{C}_{pool} = \{c^{(1)}, \dots, c^{(M)}\}$.
These descriptors include topological properties, chemical fragments, molecular properties, surface area estimates, heterocycle estimates, 3D structural properties, Lipinski parameters, and quantitative estimation of drug-likeness (QED). This step transforms the original dataset into $\mathcal{D}' = \{(x_i, y_i, \mathbf{c}^{pool}_i)\}$, providing scientifically accurate ground-truth values without manual labor.
\smallskip
\\
\noindent\textbf{Stage 2: Task-Aware Concept Selection.} 
Utilizing the entire descriptor pool $\mathcal{C}_{pool}$ ($M=200$) is suboptimal (see Figure \ref{fig:concept_ablation}) as irrelevant concepts introduce noise and dilute interpretability. To address this Relevance Gap, we employ ChatGPT (GPT-4) \cite{openai2024gpt4ocard} as a semantic filter.
Formally, let $T$ be the textual description of the downstream task (e.g., ``Predicting drug-induced liver injury''). We model the LLM as a selection operator $\mathcal{S}_{LLM}$ that identifies the indices of the top-$K$ most task-relevant concepts:
\vspace{-1mm}
\begin{equation}
    \mathcal{I}_{task} = \mathcal{S}_{LLM}(T, \mathcal{C}_{pool}) \subset \{1, \dots, M\}, \quad |\mathcal{I}_{task}| = K
\end{equation}
The final task-specific concept vector $\mathbf{c}^*_i \in \mathbb{R}^K$ for molecule $x_i$ is then constructed by gathering the selected descriptors: $\mathbf{c}^*_i = [c^{(j)}_i]_{j \in \mathcal{I}_{task}}$. 
This process yields a compact and scientifically grounded supervision target.

\vspace{-4mm}
\subsection{The \method Architecture}
\label{subsec:arch}
\vspace{-2mm}

The architecture of \method progressively transforms molecular data from raw inputs to latent embeddings, then to interpretable concept scores, and finally to task predictions. As depicted in Figure \ref{fig:1}, this process involves three tightly integrated modules.
\smallskip
\vspace{.1mm}
\noindent\textit{\textbf{Latent Feature Extraction.}} The process begins with the \textbf{Backbone Encoder} $f_\theta$, which serves as the feature extraction engine. A key design principle of \method is its adaptability to different molecular modalities. For graph-structured inputs, we instantiate $f_\theta$ with a GINEConv-based GNN \cite{hu2020strategiespretraininggraphneural} to capture topological nuances. For sequential SMILES inputs, we employ the chemistry-specific SMILY-APE LLM \cite{SMILY-APE}. Regardless of the backbone choice, this module maps the raw molecule $x$ into a latent embedding $\mathbf{e} = f_\theta(x) \in \mathbb{R}^d$.
\smallskip
\\
\noindent\textit{\textbf{Concept Projection.}} To bridge the gap between these abstract representations and human understanding, a \textbf{Concept Projector} $g_\phi$ transforms the embedding $\mathbf{e}$. Unlike standard classifiers that map embeddings directly to labels, this module—a Multi-Layer Perceptron—projects the latent vector onto the task-specific concept space defined in the previous section, yielding a predicted concept vector:
\vspace{-1mm}
\begin{equation}
    \hat{\mathbf{c}} = g_\phi(\mathbf{e}) \in \mathbb{R}^K
\end{equation}
Each element of $\hat{\mathbf{c}}$ represents the model's estimation of a physicochemical property (e.g., LogP), requiring the network to explicitly represent the molecule's attributes.
\smallskip
\\
\noindent\textit{\textbf{Transparent Inference.}} Finally, the prediction is made by the \textbf{Linear Predictor} $h_\psi$. By constraining this layer to a simple linear transformation, we ensure that the decision logic remains fully transparent:
\vspace{-1mm}
\begin{equation}
    \hat{y} = h_\psi(\hat{\mathbf{c}}) = \sigma(\mathbf{w}^\top \hat{\mathbf{c}} + b)
\end{equation}
This design guarantees that the final prediction is solely a function of the predicted concepts, with interpretability arising from the linearity of $h_\psi$.

\vspace{-4mm}
\subsection{Optimization and Interpretable Inference}
\label{subsec:opt}
\vspace{-2mm}

\noindent\textit{\textbf{Joint Optimization.}} To ensure that \method achieves both high predictive accuracy and faithful concept learning, we optimize the entire framework end-to-end using a composite objective function. The loss is formulated as:
\vspace{-1mm}
\begin{equation}
    \mathcal{L} = \mathcal{L}_{\text{task}}(y, \hat{y}) + \lambda \cdot \mathcal{L}_{\text{concept}}(\mathbf{c}^*, \hat{\mathbf{c}})
\end{equation}
where $\mathcal{L}_{\text{task}}$ is the binary cross-entropy loss for the classification target. The concept supervision term, $\mathcal{L}_{\text{concept}}$, is defined as the Mean Absolute Error (L1 loss) between the predicted concept scores $\hat{\mathbf{c}}$ and the RDKit-computed ground truth $\mathbf{c}^*$:

\vspace{-1mm}
\begin{equation}
    \mathcal{L}_{\text{concept}} = \frac{1}{K} \sum_{k=1}^K | c^*_k - \hat{c}_k |
\end{equation}
The hyperparameter $\lambda$ governs the trade-off between task performance and concept alignment. As demonstrated in our ablation study (see Figure \ref{fig:hyperparams}), setting $\lambda=1$ typically yields the optimal balance. A lower $\lambda$, moreover, has been linked to worsened interpretability and lower concept accuracy \cite{leakage}.
\smallskip
\\
\noindent\textit{\textbf{Interpretable Inference.}} Once trained, the model provides more than just a prediction; it offers a granular explanation of its decision-making process. Due to the linearity of the final predictor $h_\psi$, the logit of the prediction can be exactly decomposed into the sum of individual concept contributions. For a given molecule, the contribution score $s_k$ of the $k$-th concept is calculated as:
\vspace{-1mm}
\begin{equation}
    s_k = w_k \cdot \hat{c}_k
\end{equation}
Here, $\mathbf{w} \in \mathbb{R}^K$ directly quantifies the contribution of each concept: a large positive $s_k$ implies that concept $k$ strongly drives the model toward a positive prediction, while a negative value suggests an inhibitory effect. This enables domain experts to verify whether the model's reasoning aligns with established chemical principles.
\vspace{-4mm}
\section{Experiments}
\label{sec:exp}
\vspace{-2mm}

We conduct comprehensive experiments to validate the effectiveness of \method. These experiments aim to investigate the following research questions:
\vspace{-2mm}
\begin{itemize}
    \item \textbf{RQ1 (Effectiveness):} Can the model demonstrate competitive performance with both GNN and LLM backbones?
    \item \textbf{RQ2 (Interpretability):} Does \method learn chemically meaningful concepts?
    \item \textbf{RQ3 (Ablation):} How critical is each model component to performance?
\end{itemize}
\vspace{-2mm}

\definecolor{glassblue}{HTML}{E6F2FF}
\definecolor{glassred}{HTML}{FFF0F0}

\begin{table*}[t]
	\caption{Experimental results on Therapeutics Data Commons datasets. Besides the absence of a concept loss and concept layer, the baseline architectures are identical to their respective \method architectures. We report the AUROC (mean $\pm$ std) over five runs. \textbf{Bold} indicates the best performance within the same architecture. The columns for our method (\textbf{\method}) are highlighted.}
	\setlength{\tabcolsep}{6pt}
	\renewcommand{\arraystretch}{0.9}
	\centering

	\begin{tabular}{l c >{\columncolor{glassblue}}c @{\hspace{1.0em}} c c >{\columncolor{glassred}}c}
	\toprule
	\multicolumn{1}{c}{\multirow{2}{*}{\textbf{Dataset}}} & \multicolumn{2}{c}{\textbf{LLM Architecture}} & & \multicolumn{2}{c}{\textbf{GNN Architecture}} \\
	\cmidrule(l{14pt}r){2-3} \cmidrule(l{14pt}r){5-6}
	 & Baseline & \textbf{\method (Ours)} & & Baseline & \textbf{\method (Ours)} \\ 
	\midrule
	& \multicolumn{5}{c}{\textit{\textbf{---ADME Properties---}}} \\
	BBB & 0.856 {\tiny\textcolor{gray}{$\pm$ 0.010}} & \textbf{0.892} {\tiny\textcolor{gray}{$\pm$ 0.023}} & & 0.840 {\tiny\textcolor{gray}{$\pm$ 0.006}} & \textbf{0.867} {\tiny\textcolor{gray}{$\pm$ 0.014}} \\
	Bioavailability & 0.556 {\tiny\textcolor{gray}{$\pm$ 0.054}} & \textbf{0.679} {\tiny\textcolor{gray}{$\pm$ 0.010}} & & \textbf{0.615} {\tiny\textcolor{gray}{$\pm$ 0.044}} & 0.572 {\tiny\textcolor{gray}{$\pm$ 0.045}} \\ 
	HIA & 0.880 {\tiny\textcolor{gray}{$\pm$ 0.111}} & \textbf{0.989} {\tiny\textcolor{gray}{$\pm$ 0.007}} & & 0.768 {\tiny\textcolor{gray}{$\pm$ 0.041}} & \textbf{0.867} {\tiny\textcolor{gray}{$\pm$ 0.057}} \\ 
	Pgp & 0.801 {\tiny\textcolor{gray}{$\pm$ 0.005}} & \textbf{0.903} {\tiny\textcolor{gray}{$\pm$ 0.003}} & & \textbf{0.899} {\tiny\textcolor{gray}{$\pm$ 0.011}} & 0.861 {\tiny\textcolor{gray}{$\pm$ 0.014}} \\ 
	CYP2C9 & 0.822 {\tiny\textcolor{gray}{$\pm$ 0.001}} & \textbf{0.859} {\tiny\textcolor{gray}{$\pm$ 0.010}} & & 0.851 {\tiny\textcolor{gray}{$\pm$ 0.008}} & \textbf{0.875} {\tiny\textcolor{gray}{$\pm$ 0.004}} \\
	CYP2D6 & 0.761 {\tiny\textcolor{gray}{$\pm$ 0.004}} & \textbf{0.793} {\tiny\textcolor{gray}{$\pm$ 0.021}} & & 0.802 {\tiny\textcolor{gray}{$\pm$ 0.003}} & \textbf{0.843} {\tiny\textcolor{gray}{$\pm$ 0.005}} \\ 
	CYP3A4 & 0.799 {\tiny\textcolor{gray}{$\pm$ 0.002}} & \textbf{0.848} {\tiny\textcolor{gray}{$\pm$ 0.002}} & & 0.840 {\tiny\textcolor{gray}{$\pm$ 0.002}} & \textbf{0.868} {\tiny\textcolor{gray}{$\pm$ 0.004}} \\
	CYP2C9\_Subs & 0.504 {\tiny\textcolor{gray}{$\pm$ 0.031}} & \textbf{0.545} {\tiny\textcolor{gray}{$\pm$ 0.024}} & & 0.599 {\tiny\textcolor{gray}{$\pm$ 0.044}} & \textbf{0.627} {\tiny\textcolor{gray}{$\pm$ 0.043}} \\
	CYP2D6\_Subs & 0.670 {\tiny\textcolor{gray}{$\pm$ 0.014}} & \textbf{0.747} {\tiny\textcolor{gray}{$\pm$ 0.024}} & & 0.665 {\tiny\textcolor{gray}{$\pm$ 0.023}} & \textbf{0.695} {\tiny\textcolor{gray}{$\pm$ 0.034}} \\ 
	CYP3A4\_Subs & 0.532 {\tiny\textcolor{gray}{$\pm$ 0.015}} & \textbf{0.548} {\tiny\textcolor{gray}{$\pm$ 0.022}} & & 0.523 {\tiny\textcolor{gray}{$\pm$ 0.036}} & \textbf{0.540} {\tiny\textcolor{gray}{$\pm$ 0.039}} \\ 
	
	\midrule
	& \multicolumn{5}{c}{\textit{\textbf{---Toxicity---}}} \\ 
	DILI & 0.814 {\tiny\textcolor{gray}{$\pm$ 0.008}} & \textbf{0.851} {\tiny\textcolor{gray}{$\pm$ 0.011}} & & 0.818 {\tiny\textcolor{gray}{$\pm$ 0.030}} & \textbf{0.850} {\tiny\textcolor{gray}{$\pm$ 0.027}} \\
	hERG & 0.726 {\tiny\textcolor{gray}{$\pm$ 0.027}} & \textbf{0.765} {\tiny\textcolor{gray}{$\pm$ 0.013}} & & \textbf{0.790} {\tiny\textcolor{gray}{$\pm$ 0.030}} & 0.695 {\tiny\textcolor{gray}{$\pm$ 0.029}} \\ 
	AMES & 0.792 {\tiny\textcolor{gray}{$\pm$ 0.009}} & \textbf{0.842} {\tiny\textcolor{gray}{$\pm$ 0.008}} & & 0.855 {\tiny\textcolor{gray}{$\pm$ 0.005}} & 0.855 {\tiny\textcolor{gray}{$\pm$ 0.003}} \\ 
	
	\midrule
	\textbf{Average} & 0.732 & \textbf{0.789} & & 0.766 & \textbf{0.778} \\
	\bottomrule
	\end{tabular}
	\label{table:2}
    \vspace{-0.6cm}
\end{table*}

\vspace{-4mm}
\subsection{Experimental Setup}
\label{subsec:setup}
\vspace{-2mm}

\paragraph{\textbf{Datasets.}} We evaluate our framework on 13 widely-used benchmark datasets from the Therapeutics Data Commons \cite{huang2021therapeuticsdatacommonsmachine}. These datasets are categorized into ADME properties (e.g., BBB, HIA) and Toxicity tasks (e.g., DILI, AMES). For all datasets, we employ a \textit{scaffold split} (80/10/10) to rigorously evaluate the model's generalization capability to unseen chemical structures.
\vspace{-0.3cm}
\paragraph{\textbf{Baselines.}} To evaluate the effectiveness of our proposed framework, we compare it against baseline methods from two primary categories: (1) \textit{GNN Baselines}: We include established GNNs such as GINE \cite{hu2020strategiespretraininggraphneural}, GCN, GAT, and GraphSAGE; (2) \textit{LLM Baselines}: We benchmark against various LLM-based methods, primarily the chemistry-specific SMILY-APE \cite{SMILY-APE}, as well as Qwen and Llama.
\vspace{-0.3cm}
\paragraph{\textbf{Implementation Details.}} For the bottleneck layer, we utilize RDKit to generate ground-truth concepts and ChatGPT for task-aware selection. Following the guidance of our ablation study, the hyperparameter $\lambda$ is set to 1, and 40 concepts are chosen. All experiments and analyses are conducted on a single NVIDIA H100 GPU.

\vspace{-4mm}
\subsection{Performance Comparison (RQ1)}
\label{subsec:rq1}
\vspace{-2mm}

Table \ref{table:2} compares \method against black-box baselines to examine the relationship between transparency and performance.

\vspace{-2mm}
\paragraph{\textbf{Challenging the Trade-Off.}} Contrary to common concerns, our results demonstrate that interpretability does not necessitate a compromise in performance. In the LLM setting, \method consistently outperforms the baseline across all datasets, achieving an average AUROC improvement of 0.057. Similarly, in the GNN setting, our interpretable model outperforms the black-box baselines on 9 out of 13 tasks and ties on one, with an average improvement of 0.012.

\vspace{-3mm}
\paragraph{\textbf{Task-Specific Gains.}} Notably, the performance gains are most pronounced in toxicity tasks (e.g., DILI, AMES). This suggests that the explicit modeling of chemical concepts helps the model focus on specific toxicophores (structural alerts) rather than overfitting to spurious correlations in the high-dimensional latent space. 
This addresses \textbf{RQ1}: \method demonstrates that the CBM architecture performs competitively with baseline models, with the LLM architecture showing consistent improvements and the GNN architecture remaining competitive on most tasks.
\begin{figure}[t!]
    \centering
    \begin{subfigure}[b]{0.40\textwidth}
        \centering
        \includegraphics[width=\textwidth]{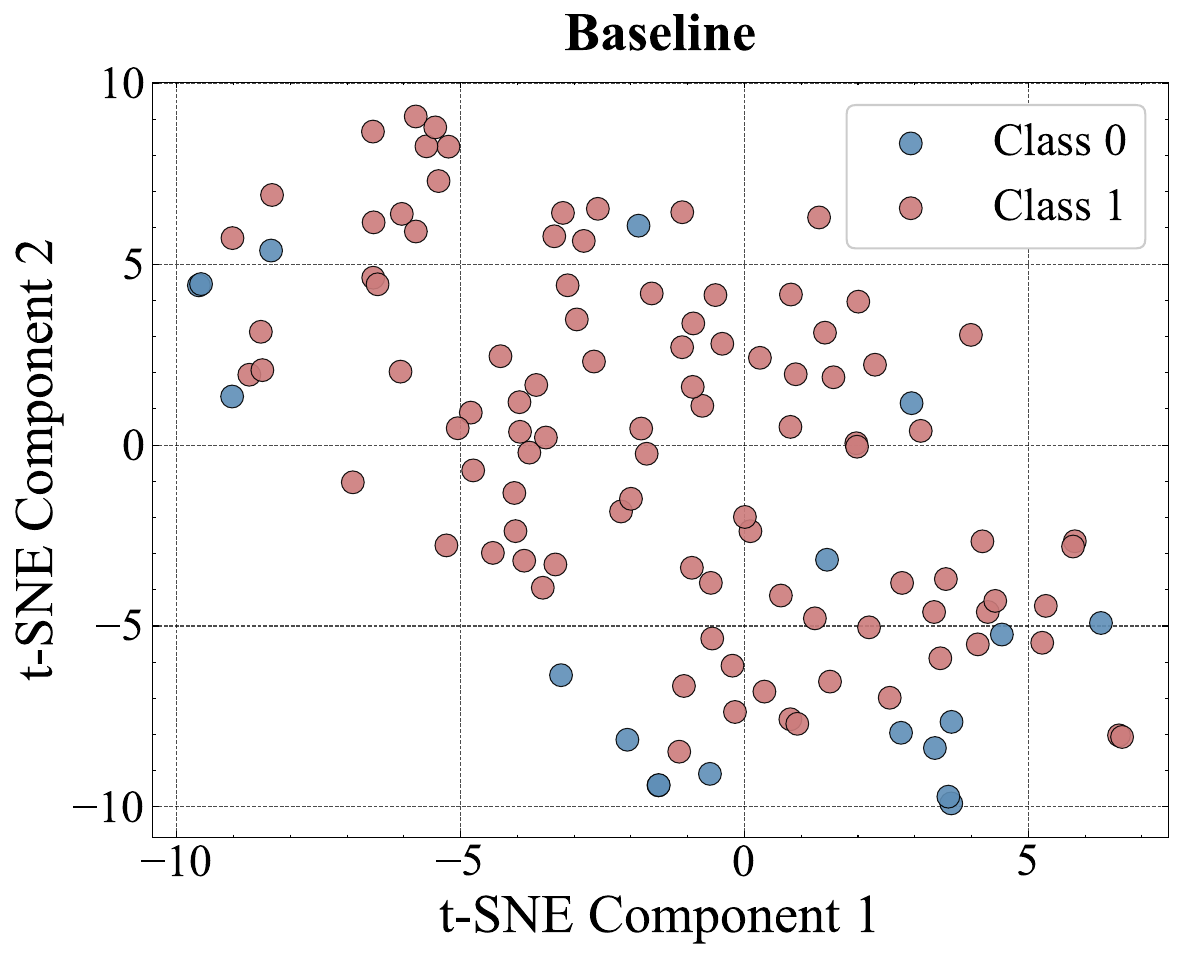}
        \label{fig:tsne_baseline}
    \end{subfigure}
    \hspace{4mm}
    \begin{subfigure}[b]{0.40\textwidth}
        \centering
        \includegraphics[width=\textwidth]{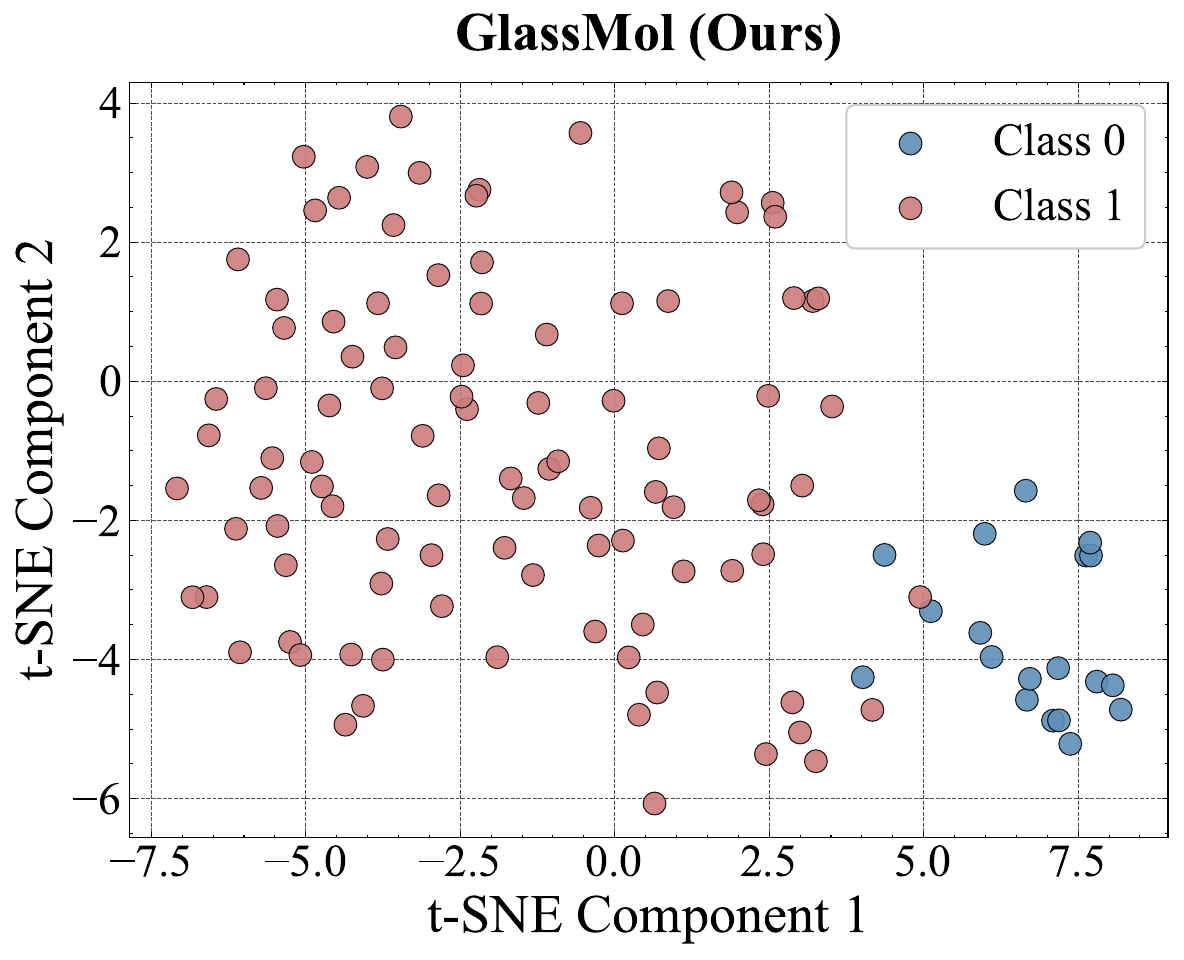}
        \label{fig:tsne_ours}
    \end{subfigure}
    \vspace{-0.7cm}
    \caption{\textbf{t-SNE visualization of learned representations on the HIA dataset.} 
    \textbf{Left:} The baseline shows entangled representations with no clear decision boundary. 
    \textbf{Right:} \method (ours) produces well-separated clusters for the two classes.}
    \label{fig:tsne_visualization}
    \vspace{-0.4cm}
\end{figure}

\begin{figure}[t!]
    \centering
    \includegraphics[page=3, scale=0.26, trim={00mm 62mm 30mm 55mm}, clip]{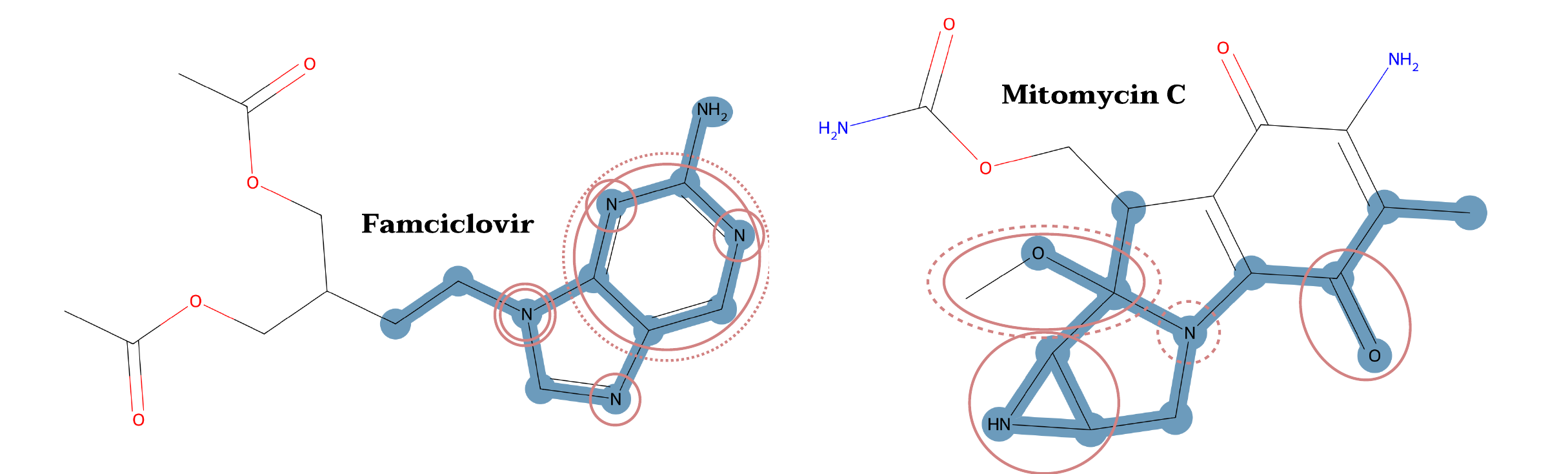}
    \vspace{-0.2cm}
    \caption{\textbf{Qualitative comparison of concept attribution against ground-truth structural importance.} The blue highlights represent the TopoPool ground-truth importance clusters. Red circles indicate the substructures identified by \method where solid lines $\xrightarrow{}$ GNN architecture and dashed lines $\xrightarrow{}$ LLM architecture.}
    \label{fig:case_study}
    \vspace{-0.7cm}
\end{figure}

\begin{figure}[t!]
    \centering
    \includegraphics[width=\textwidth]{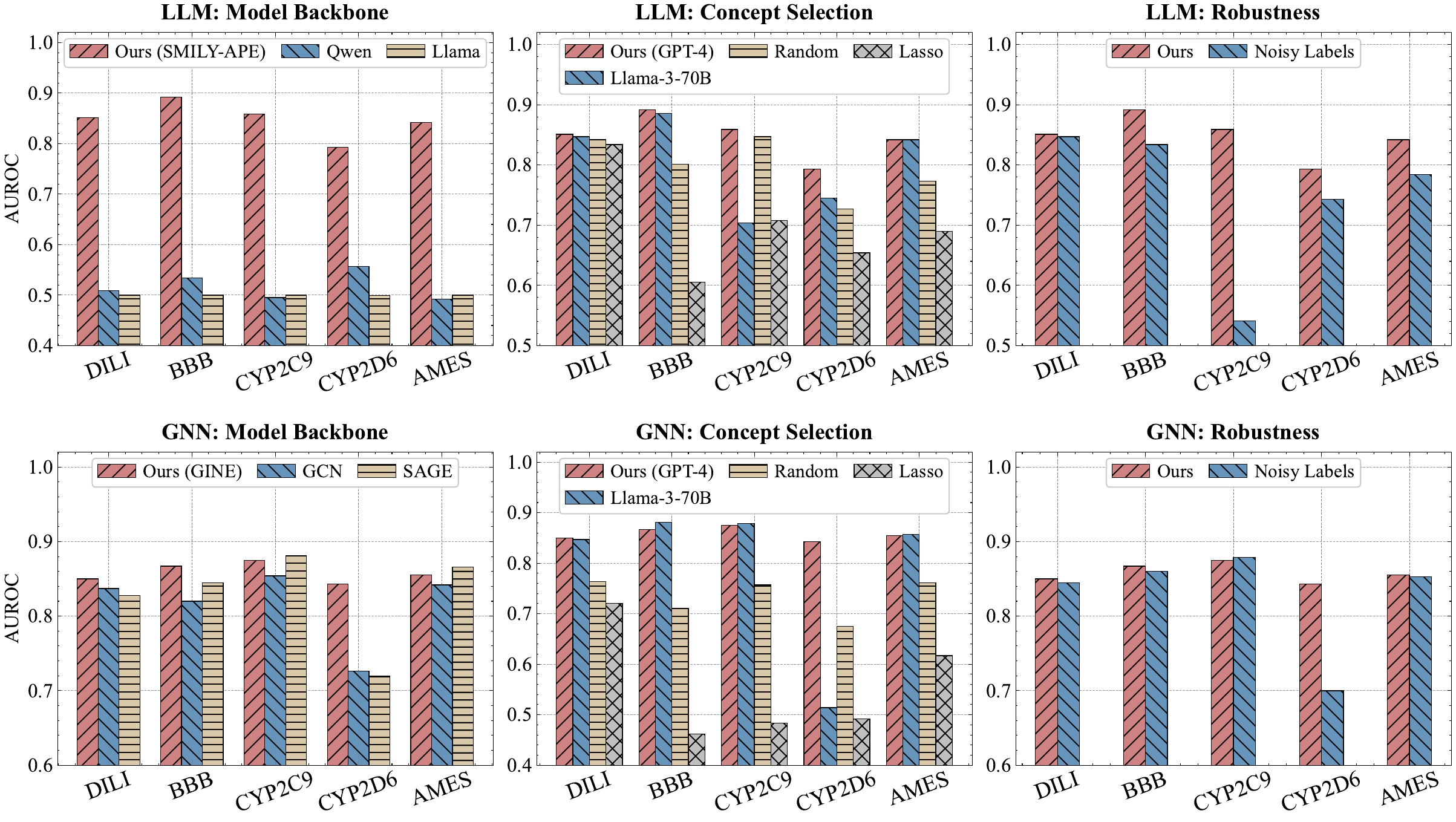}
    \vspace{-0.6cm}
    \caption{\textbf{Ablation studies on critical model components for LLM (top row) and GNN (bottom row) architectures.} 
    \textbf{Left:} Comparison of model backbones. For LLMs, our task-specific small model (red) significantly outperforms general large models (blue/beige). For GNNs, our GINE-based architecture remains competitive.
    \textbf{Middle:} Evaluation of concept selection strategies. GPT-4 selected concepts (red) achieve the best performance, with the open-source Llama-3-70B selector (blue) serving as a strong alternative to proprietary models.
    \textbf{Right:} Robustness analysis showing that our method maintains high performance even with noisy concept labels (blue).}
    \label{fig:ablation_study}
    \vspace{-0.6cm}
\end{figure}

\vspace{-4mm}
\subsection{Interpretability and Verification (RQ2)}
\label{subsec:rq2}
\vspace{-2mm}

To address \textbf{RQ2}, we validate whether the learned concepts are chemically meaningful through both latent space visualization and granular case studies.

\vspace{-4mm}
\paragraph{\textbf{Latent Space Disentanglement.}} Figure \ref{fig:tsne_visualization} visualizes the t-SNE embeddings of the penultimate layer. While the baseline model (left) produces entangled representations with no clear decision boundary, \method (right) yields distinct, well-separated clusters. This suggests that explicit concept supervision helps disentangle the latent chemical space, enabling the linear predictor to make robust, informed decisions.
\vspace{-3.5mm}
\paragraph{\textbf{Case Studies and Ground Truth Verification.}} We verify the faithfulness of our explanations by comparing the concept contributions of molecular substructures ($s_k$) against structural importance maps from TopoPool \cite{thieme2023topopool}. We select two representative molecules from different tasks to demonstrate interpretability. Note that the overlap between highlights and circles suggests strong model alignment with the ground truth. As illustrated in Figure \ref{fig:case_study}, we analyze the following molecules:
\vspace{-2mm}
\begin{itemize}
    \item \textbf{Famciclovir (HIA Task):} The GNN-based \method correctly identifies the \textit{aniline}, \textit{aromatic nitrogens}, and \textit{tertiary amines} as the largest contributors, aligning with TopoPool's assessment. Interestingly, the LLM architecture shows a more focused attribution pattern, solely highlighting the \textit{aniline} group with a contribution score ten times larger than the second factor.
    \item \textbf{Mitomycin C (DILI Task):} The GNN-based \method correctly identifies the \textit{methoxy group}, \textit{carbonyl oxygens}, and the \textit{piperazine group} as key drivers for liver injury prediction. The LLM architecture again chooses fewer concepts but remains well aligned with TopoPool. These substructures are known metabolic liabilities, confirming that the model aligns with medicinal chemistry intuition.
\end{itemize}
\vspace{-2mm}
These results address \textbf{RQ2}, confirming that \method learns faithful and chemically grounded concepts.

\begin{figure}[t]
    \centering
    \begin{subfigure}[b]{0.40\textwidth}
        \centering
        \includegraphics[width=\textwidth]{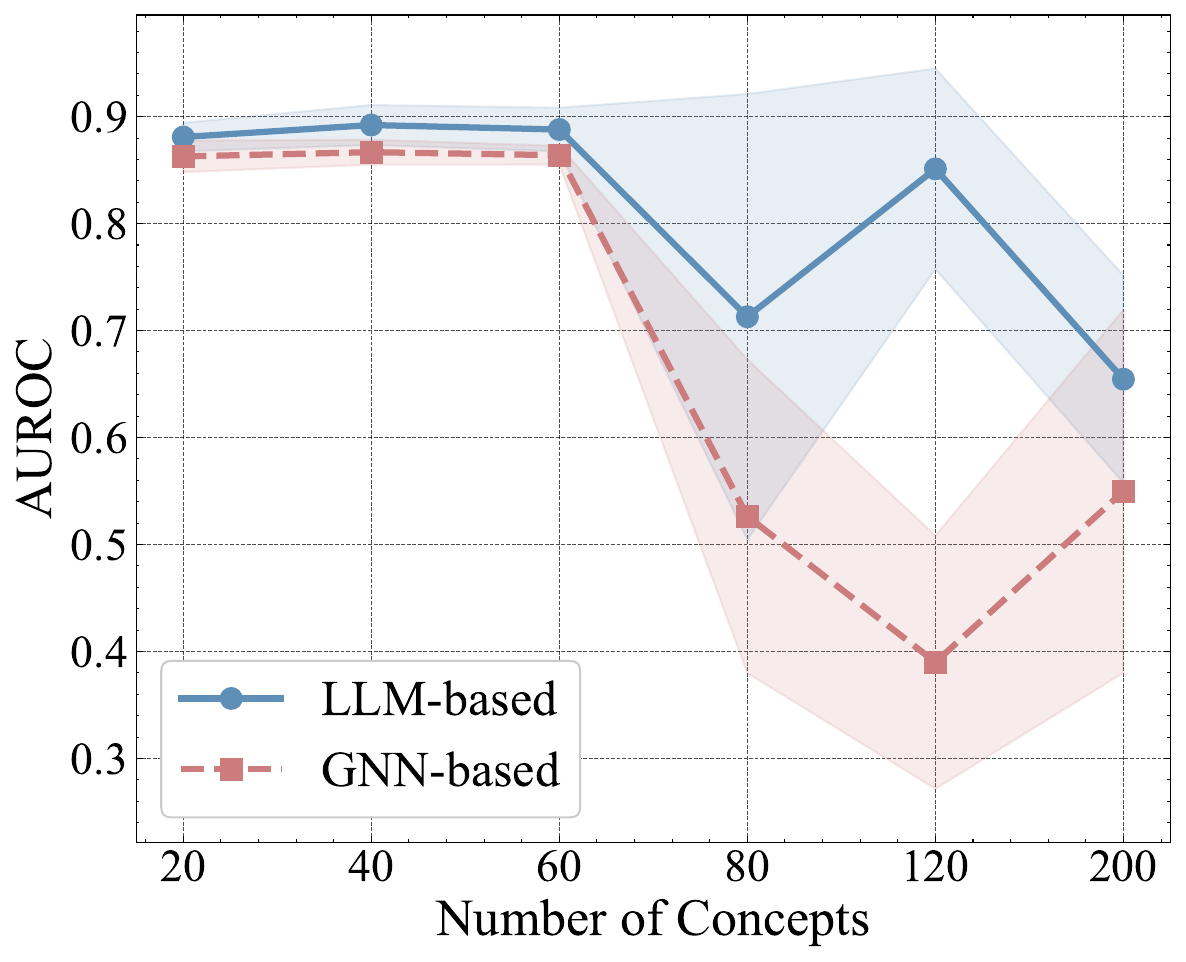}
        \caption{Impact of concept quantity ($K$)}
        \label{fig:concept_ablation}
    \end{subfigure}
    \hspace{4mm}
    \begin{subfigure}[b]{0.40\textwidth}
        \centering
        \includegraphics[width=\textwidth]{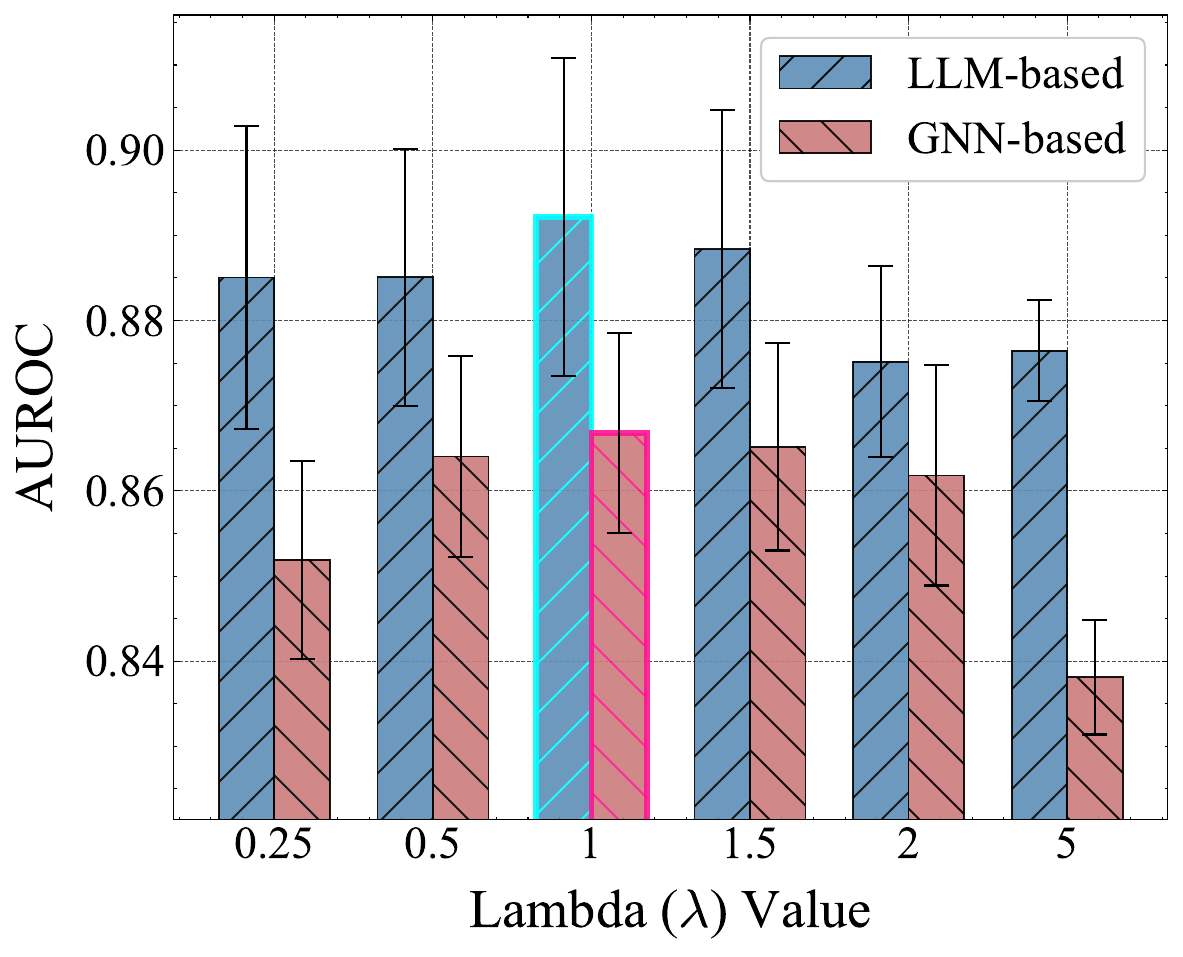}
        \caption{Sensitivity to weight parameter $\lambda$}
        \label{fig:lambda_ablation}
    \end{subfigure}
    \vspace{-0.2cm}
    \caption{\textbf{Hyperparameter sensitivity analysis.} (a) Performance vs. number of concepts ($K$). Accuracy improves rapidly and saturates around $K=40$, balancing representation power and complexity. (b) Performance vs. loss weight $\lambda$. A balanced weight of $\lambda=1$ yields optimal results, while extreme values degrade performance.}
    \label{fig:hyperparams} 
    \vspace{-.6cm}
\end{figure}

\vspace{-4mm}
\subsection{Ablation Studies (RQ3)}
\label{subsec:rq3}
\vspace{-2mm}

We investigate the contribution of individual components to answer \textbf{RQ3}. The comprehensive results are visualized in Figure \ref{fig:ablation_study}, covering three critical dimensions.

\vspace{-4mm}
\paragraph{\textbf{Model Backbone (Left Panel).}}
We investigate the influence of architecture and model size on performance.
\vspace{-1mm}
\begin{itemize}
    \item \textit{LLM:} Replacing our default chemistry-specific small model (SMILY-APE, 34.7M) with general-purpose models like Qwen-7B and Llama-8B leads to substantial performance degradation (red vs. blue/beige bars). This highlights that \textbf{domain-specific pre-training is critical} for molecular concept learning.
    \item \textit{GNN:} For graph encoders, our GINE-based architecture remains competitive. While SAGE marginally outperforms GINE on 2 datasets, GINE provides the most consistent performance across the board, justifying its selection as the default backbone.
\end{itemize}
\vspace{-1mm}

\vspace{-4mm}
\paragraph{\textbf{Concept Selection Strategy (Middle Panel).}} We probe the concept selection strategy to determine whether or not a large LLM selector is necessary.
The results show that using chemically relevant concepts selected by GPT-4 (red) significantly outperforms random selection (beige). Lasso-based sparse learning (gray) performs poorly, suggesting that end-to-end concept discovery struggles to converge.
Notably, the open-source Llama-3-70B selector (blue) achieves performance comparable to GPT-4. This is a crucial finding for practicality, allowing \method to run entirely on local hardware.

\vspace{-3mm}
\paragraph{\textbf{Robustness (Right Panel).}} 
Real-world data is often imperfect. We test \method with noisy labels by randomly perturbing ground-truth concepts. Remarkably, the performance drop is negligible (red vs. blue bars), demonstrating that the architecture is robust to annotation noise. This suggests that the model learns to extract meaningful patterns from imperfect supervision rather than memorizing exact concept values.

\vspace{-4mm}
\paragraph{\textbf{Hyperparameter Sensitivity.}}
We analyze the impact of two key hyperparameters: the number of concepts $K$ and the loss balancing weight $\lambda$. 
Figure \ref{fig:concept_ablation} shows that model performance improves significantly as $K$ increases from 10 to 40, indicating that a sufficient number of concepts is required to capture complex chemical properties. However, the performance plateau beyond $K=40$ suggests that adding more concepts yields diminishing returns and may introduce noise.
Regarding the loss weight, Figure \ref{fig:lambda_ablation} demonstrates that $\lambda=1$ offers the optimal trade-off. A small $\lambda$ fails to align the concept space (poor interpretability), while an excessively large $\lambda$ forces the model to prioritize concept reconstruction over the primary classification task.
\vspace{-4mm}
\section{Conclusion}
\vspace{-2mm}
In this work, we apply CBMs to molecular property prediction, allowing a model to project GNN and LLM embeddings to the concept space before generating a prediction. By leveraging RDKit for automated concept computation and LLMs for task-aware concept selection, we effectively bridge the annotation and relevance gaps that have previously hindered the adoption of CBMs for molecular tasks. Our experiments on thirteen diverse chemical datasets demonstrate that \method generally matches or exceeds black-box baselines, suggesting that interpretability need not come at the cost of performance. Case studies further validate that the learned concept attributions align with established structural importance methods. As machine learning methods evolve in both quality and complexity, the CBM architecture offers a promising path to keep human experts informed and engaged in the prediction process, potentially accelerating the discovery of safe and effective therapeutics.

\vspace{-2mm}
\begin{credits}
\subsubsection{\ackname} This research was supported in part through the computational resources and staff contributions provided for the Quest high performance computing facility at Northwestern University which is jointly supported by the Office of the Provost, the Office for Research, and Northwestern University Information Technology. Funding for this research was provided by the OURG at Northwestern University in the form of a SURG grant. 

\vspace{-2mm}
\subsubsection{\discintname}
Kaize Ding is the head of and Oscar Rivera and Ziqing Wang are members of the REAL Lab which is participating in a collaborative research effort between Northwestern University and AbbVie Inc.
\end{credits}

\vspace{-4mm}
\bibliographystyle{splncs04}
\bibliography{pakdd}
%
%
%
%
%
%
%
\end{document}